\documentclass[11pt,a4paper]{article}
\usepackage[hyperref]{acl2018}
\usepackage{times}
\usepackage{latexsym}

\usepackage{url}

\usepackage{figsize}
\usepackage[normalem]{ulem}
\usepackage{xspace}
\usepackage{amsmath}
\usepackage{amssymb}
\usepackage{arydshln}
\usepackage{enumitem}
\usepackage{url}
\usepackage{soul}
\usepackage{xcolor}
\usepackage[ruled,noline,linesnumbered]{algorithm2e}
\usepackage{pifont}
\usepackage[font=small]{caption}

\setlist{nolistsep}

\newcommand{\Note}[1]{}
\renewcommand{\Note}[1]{\hl{[#1]}}  
\definecolor{brilliantlavender}{rgb}{0.96, 0.73, 1.0}
\definecolor{brightgreen}{rgb}{0.4, 1.0, 0.0}
\definecolor{mikadoyellow}{rgb}{1.0, 0.77, 0.05}

\newcommand{\direct}{DA\xspace}
\newcommand{\rank}{RA\xspace}

\newcommand{\method}{EASL\xspace}
\newcommand{\methodfull}{{\bf E}fficient {\bf A}nnotation of {\bf S}calar {\bf L}abels\xspace}

\newcommand{\object}{S_i}
\newcommand{\thur}{Thurstone\xspace}

\newcommand{\ts}{TrueSkill\xspace}

\newcommand{\N}{\mathcal{N}}
\newcommand{\B}{\mathcal{B}}

\newcommand{\mm}{\mathbb{M}}
\newcommand{\vwin}{v_{i \succ j}}
\newcommand{\vdraw}{v_{i \equiv j}}
\newcommand{\wwin}{w_{i \succ j}}
\newcommand{\wdraw}{w_{i \equiv j}}

\usepackage{todonotes} 

\aclfinalcopy 
\def\aclpaperid{1300} 

\newcommand{\repo}{\url{http://decomp.net/}}

\title{Efficient Online Scalar Annotation with Bounded Support}
\author{
  Keisuke Sakaguchi
 \and Benjamin Van Durme\\
 Johns Hopkins University\\
 {\tt \{keisuke,vandurme\}@cs.jhu.edu}
}

\date{}

\begin{document}
\maketitle
\begin{abstract}
We describe a novel method for efficiently eliciting scalar annotations for dataset construction and system quality estimation by human judgments. 
We contrast direct assessment (annotators assign scores to items directly), online pairwise ranking aggregation (scores derive from annotator comparison of items), and a hybrid approach (\method: \methodfull) proposed here.  
Our proposal leads to increased correlation with ground truth, at far greater annotator efficiency, suggesting this strategy as an improved mechanism for dataset creation and manual system evaluation.
\end{abstract}

\section{Introduction}
\vspace{-2mm}
We are concerned here with the construction of datasets and evaluation of systems within natural language processing (NLP).  Specifically, humans providing responses that are used to derive graded values on natural language contexts, or in the ordering of systems corresponding to their perceived performance on some task.

\begin{figure}[t]
	\centering
    \vspace{-2mm}
	\includegraphics[width=77mm]{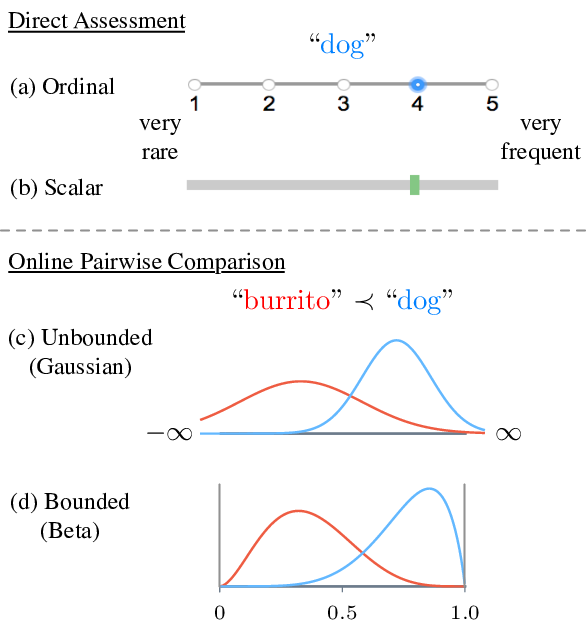}
	\caption{Elicitation strategies for graded response include direct assessment via ordinal or scalar judgments, and pairwise comparisons aggregated via an assumption of latent distributions such as Gaussians, or novel here:  Beta distributions, providing bounded support. The example concerns subjective assessments of the lexical frequency of \emph{dog}. In pairwise comparison, we assess it by comparison such as ``burrito'' is less frequent ($\prec$) than ``dog''.}
    \label{fig:highlevel}
	\vspace{-2mm}
\end{figure}
Many NLP datasets involve eliciting from annotators some graded response.
The most popular annotation scheme is the $n$-ary ordinal approach as illustrated in Figure~\ref{fig:highlevel}(a).
For example, text may be labeled for \emph{sentiment} as \emph{positive}, \emph{neutral} or \emph{negative} \cite[inter alia]{wiebe1999development,pang2002thumbs,turney2002thumbs}; or  under \emph{political spectrum analysis} as \emph{liberal}, \emph{neutral}, or \emph{conservative}~\cite{o2010tweets,bamman-smith:2015:EMNLP}.  A response may correspond to a likelihood judgment, e.g., how likely a predicate is factive~\cite{lee-EtAl:2015:EMNLP}, or that some natural language inference may hold~\cite{TACL1082}.  Responses may correspond to a notion of semantic similarity, e.g., whether one word can be substituted for another in context \cite{pavlick-EtAl:2015:ACL-IJCNLP2}, or whether an entire sentence is more or less similar than another \cite{L14-1314}, and so on.

Less common in NLP are system comparisons based on direct human ratings, but an exception includes the annual shared task evaluations of the Conference on Machine Translation (WMT).  There, MT practitioners submit system outputs based on a shared set of source sentences, which are then judged relative to other system outputs.  Various aggregation strategies have been employed over the years to take these relative comparisons and derive competitive rankings between shared task entrants~\cite{callisonburch-EtAl:2012:WMT,bojar-EtAl:2013:WMT,bojar-EtAl:2014:W14-33,bojar-EtAl:2015:WMT,bojar-EtAl:2016:WMT1,bojar-EtAl:2017:WMT1}.

Inspired by prior work in MT system evaluation, we propose a procedure for eliciting graded responses that we demonstrate to be more efficient than prior work.
While remaining applicable to system evaluation, our experimental results suggest our approach as a more general framework for a variety of future data creation tasks, allowing for higher quality data in less time and cost.

We consider three different approaches for scalar annotation: direct assessment (\direct), online pairwise ranking aggregation (\rank), and a hybrid method which we call \method (\methodfull).\footnote{Pronounced as ``easel''.}
\direct scalar annotation, shown in Figure~\ref{fig:highlevel}(b), directly annotates absolute judgments on some scale (e.g., 0 to 100), independently per item (\S\ref{sec:direct}).
As an \rank approach (\S\ref{sec:rank}), we start with conventional unbounded models, where each instance is parameterized as a Gaussian distribution, as shown in Figure~\ref{fig:highlevel}(c).
Since boundedness is essential for the scalar annotation we aim to model, we propose a bounded variant which parameterizes each instance by a beta distribution as illustrated in Figure~\ref{fig:highlevel}(d).
Finally, we propose \method (\S\ref{sec:method}) that combines benefits of \direct and \rank.

We illustrate the improvements enabled by our proposal on three example tasks (\S\ref{sec:experiments}): lexical frequency inference, political spectrum inference and  machine translation system ranking.\footnote{We release the code at \repo.}  For example, we find that in the commonly employed condition of 3-way redundant annotation, our approach on multiple tasks gives similar quality with just 2-way redundancy: this  translates to a potential 50\% increase in dataset size for the same cost.


\section{Direct Assessment}
\vspace{-2mm}
\label{sec:direct}
Direct assessment or direct annotation (\direct) is a straightforward method for collecting graded response from annotators.
The most popular scheme is $n$-ary ordinal labeling, as illustrated in Figure~\ref{fig:highlevel}(a), where annotators are shown one instance (i.e., sample point) and asked to label one of the $n$-ary ordered classes.

According to the level of measurement in psychometrics~\cite[inter alia]{Stevens677}, which classifies the numerals based on certain properties (e.g., identity, order, quantity), ordinal data do not allow for degree of difference.
Namely, there is no guarantee that the distance between each label is equal, and instances in the same class are not discriminated.
For example, in a typical five-level Likert scale~\cite{likert1932technique} of likelihood --  very unlikely,  unlikely, unsure, likely, very likely -- we cannot conclude that \emph{very likely} instances are exactly twice as likely those marked \emph{likely}, nor can we assume two instances with the same label have exactly the same likelihood.

The issue of distance between ordinals is perhaps obviated by using \emph{scalar} annotations (i.e., \emph{ratio scale} in Stevens's terminology), which directly correspond to continuous quantities (Figure~\ref{fig:highlevel}(b)).   In scalar \direct,\footnote{In the rest of the paper, we take \direct to mean \emph{scalar} annotation rather than ordinals.} each instance in the collection ($S_i \in S_1^N$) is annotated with values (e.g., on the range 0 to 100) often by several annotators.
The notion of quantitative difference is enabled by the property of \emph{absolute} zero: the scale is \emph{bounded}.
For example, distance, length, mass, size etc. are represented by this scale.
In the annual shared task evaluation of the WMT,  \direct has been used for scoring adequacy and fluency of machine learning system outputs with human evaluation~\cite{graham-EtAl:2013:LAW7-ID,graham-EtAl:2014:EACL,bojar-EtAl:2016:WMT1,bojar-EtAl:2017:WMT1}, and has separately been used in creating datasets such as for factuality~\cite{lee-EtAl:2015:EMNLP}.

Why  \emph{perhaps} obviated?  Because of two concerns: (1) annotators may not have a pre-existing, well-calibrated scale for performing DA on a particular collection according to a particular task;\footnote{E.g., try to imagine your level of  calibration to a hypothetical task described as "On a scale of 1 to 100, label this tweet according to a conservative / liberal political spectrum."} and (2) it is known that people may be biased in their scalar estimates~\cite{Tversky1124}.  Regarding (1), this motivates us to consider RA on the intuition that annotators may give more calibrated responses when performed in the context of other elements.  Regarding (2), our goal is not to correct for human bias, but simply to  more efficiently converge to the same consensus judgments already being pursued by the community in their annotation protocols, biased or otherwise.\footnote{There has been a line of work on relative weighting of \emph{annotators}, based on their agreement with others~\cite{whitehill2009whose,welinder2010multidimensional,hovy-EtAl:2013:NAACL-HLT}. In this paper, however, we do not perform such annotator weighting.}

\section{Online Pairwise Ranking Aggregation} 
\label{sec:rank}
\subsection{Unbounded Model} 
\label{sec:unbounded}
\vspace{-0mm}
Pairwise ranking aggregation~\cite{thurstone1927method} is a method to obtain a total ranking on instances, assuming that scalar value for each sample point follows a Gaussian distribution, $\N(\mu_i, \sigma^2)$.
The parameters \{$\mu_i$\} are interpreted as  mean scalar annotation.\footnote{\thur and another popular ranking method by \newcite{elo1978rating} use a fixed $\sigma$ for all instances.}

Given the parameters, the probability that $S_i$ is preferred ($\succ$) over $S_j$ is defined as
\vspace{-0mm}
\begin{align}
p(S_i \succ S_j) &= \Phi\left( \frac{\mu_i - \mu_j}{\sqrt{2}\sigma} \right), \label{eq:thurstone}
\end{align}
where $\Phi(\cdot)$ is the cumulative distribution function of the standard normal distribution.
The objective of pairwise ranking aggregation (including all the following models) is formulated as a maximum log-likelihood estimation:
\vspace{-0mm}
\begin{align}
\max_{\{S_1^N \}} \sum_{S_i, S_j \in \{S_1^N\} } \log p(S_i \succ S_j).
\end{align}

TrueSkill\textsuperscript{TM}~\cite{HerbrichMG06} extends the \thur model by applying a Bayesian online and active learning framework,  allowing for ties. 
\ts has been used in the Xbox Live online gaming community,\footnote{\url{www.xbox.com/live/}} and 
has been applied for various NLP tasks, such as question difficulty estimation~\cite{liu-EtAl:2013:EMNLP}, ranking speech quality~\cite{baumann2017large}, and ranking machine translation and grammatical error correction systems with human evaluation~\cite{bojar-EtAl:2014:W14-33,bojar-EtAl:2015:WMT,sakaguchi-post-vandurme:2014:W14-33,TACL800}
\begin{figure}[t]
  \centering
  \SetFigLayout{2}{2}
    \subfigure[$\vwin$]{\includegraphics[width=38mm]{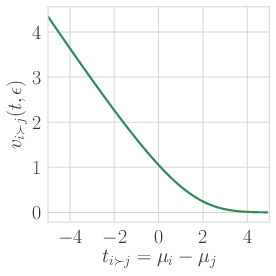}}
    \hfill
    \subfigure[$\vdraw$]{\includegraphics[width=38mm]{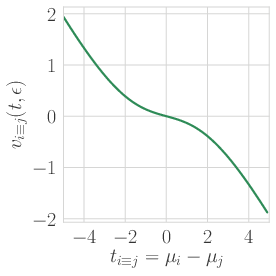}}
    \hfill
    \subfigure[$\wwin$]{\includegraphics[width=38mm]{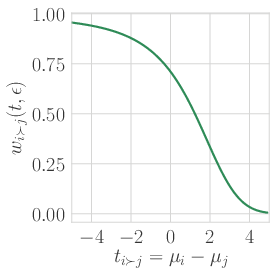}}
    \hfill
    \subfigure[$\wdraw$]{\includegraphics[width=38mm]{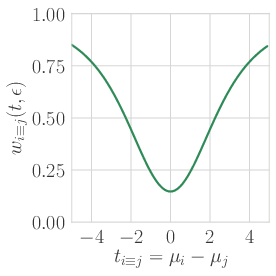}}
	\caption{Surprisal of the outcome for $\mu$ and $\sigma^2$ ($\epsilon=0.5)$.}
    \label{fig:update}
	\vspace{-0mm}
\end{figure}

In the same way as the \thur model, \ts assumes that scalar values for each instance $S_i$ (i.e., skill level for each player in the context of \ts) follow a Gaussian distribution $\N(\mu_i, \sigma^2_i)$, where $\sigma_i$ is also parameterized as the \emph{uncertainty} of the scalar value for each instance.
Importantly, \ts uses a Bayesian online learning scheme, and the parameters are {\em iteratively} updated after each observation of pairwise comparison (i.e., game result: win ($\succ$), tie ($\equiv$), or loss ($\prec$)) in proportion to how surprising the outcome is.
Let $t_{i\succ j}=\mu_{i} - \mu_{j}$, the difference in scalar responses (skill levels) when we observe $i$ wins $j$, and $\epsilon \geqslant 0$ be a parameter to specify the tie rate. 
The update functions are formulated as follows:
\vspace{-0mm}
\begin{align}
    \mu_i &= \mu_i + \frac{\sigma^2_i}{c} \cdot v\left(\frac{t}{c}, \frac{\epsilon}{c}\right) \label{eq:tsupdate1} \\
    \mu_j &= \mu_j - \frac{\sigma^2_j}{c} \cdot v\left(\frac{t}{c}, \frac{\epsilon}{c}\right) \label{eq:tsupdate2},
\end{align}
where $c^2 = 2\gamma^2 + \sigma^2_i + \sigma^2_j$, and $v$ are multiplicative factors that affect the amount of change (surprisal of the outcome) in $\mu$.
In the accumulation of the variances ($c^2$), another free parameter called ``skill chain'', $\gamma$, indicates the width (or difference) of skill levels that two given players have 0.8 (80\%) probability of win/lose.
The multiplicative factor depends on the observation (wins or ties):
\begin{align}
  \vwin(t,\epsilon) &= \frac{\varphi(-\epsilon + t)}{\Phi(-\epsilon + t)} \label{eq:tsloss1}, \\
  \vdraw(t,\epsilon) &= \frac{\varphi(-\epsilon - t) - \varphi(\epsilon - t)}{\Phi(\epsilon-t) - \Phi(-\epsilon-t)}, \label{eq:tsloss2}
\end{align}
where $\varphi(\cdot)$ is the probability density function of the standard normal distribution.
As shown in Figure~\ref{fig:update} (a) and (b), $\vwin$ increases exponentially as $t$ becomes smaller (i.e., the observation is unexpected), whereas $\vdraw$ becomes close to zero when $|t|$ is close to zero.
In short, $v$ becomes larger as the outcome is more surprising.

In order to update variance ($\sigma^2$), another set of update functions is used:
\begin{align}
  \sigma^2_i &= \sigma^2_i \cdot \left[ 1-\frac{\sigma^2_i}{c^2} \cdot w\left(\frac{t}{c}, \frac{\epsilon}{c}\right)\right] \\
  \sigma^2_j &= \sigma^2_j \cdot \left[ 1-\frac{\sigma^2_j}{c^2} \cdot w\left(\frac{t}{c}, \frac{\epsilon}{c}\right)\right],
\end{align}
where $w$ serve as multiplicative factors that affect the amount of change in $\sigma^2$.
\begingroup\makeatletter\def\f@size{9}\check@mathfonts
\def\maketag@@@#1{\hbox{\m@th\large\normalfont#1}}%
\begin{align}
    \wwin(t,\epsilon) &= \vwin \cdot \left( \vwin + t -\epsilon \right) \\
    \wdraw(t, \epsilon) &=  \vdraw^2 + \frac{(\epsilon-t)\cdot \varphi(\epsilon - t) + (\epsilon+t) \cdot \varphi(\epsilon + t)}{ \Phi(\epsilon-t) - \Phi(-\epsilon-t)}.
\end{align}\endgroup
As shown in Figure~\ref{fig:update} (c) and (d), the value of $w$ is between 0 and 1.
The underlying idea for the variance updates is that these updates always decrease the size of the variances $\sigma^2$, which means uncertainty of the instances ($S_i, S_j$) always decreases as we observe more pairwise comparisons.
In other words, \ts becomes more confident in the current estimate of $\mu_i$ and $\mu_j$.
Further details are provided by \newcite{HerbrichMG06}.\footnote{The following material is also useful to understand the math behind \ts (\url{http://www.moserware.com/assets/computing-your-skill/The\%20Math\%20Behind\%20TrueSkill.pdf}).}

Another important property of \ts is ``match quality (chance to draw)''.
The match quality helps selecting competitive players to make games more interesting.
More broadly, the match quality enables us to choose similar instances to be compared to maximize the information gain from pairwise comparisons, as in the active learning literature \cite{settles2008active}.
The match quality between two instances (players) is computed as follows:
\begingroup\makeatletter\def\f@size{10}\check@mathfonts
\def\maketag@@@#1{\hbox{\m@th\large\normalfont#1}}%
\vspace{-0mm}
\begin{align}
q(\gamma, S_i, S_j) :=  \sqrt[]{\frac{2\gamma^2}{c^2}} \exp{\left(- \frac{(\mu_i-\mu_j)^2}{2 c^2}  \right) } \label{eq:tsmatch} 
\end{align}
\endgroup
Intuitively, the match quality is based on the difference $\mu_i-\mu_j$. 
As the difference becomes smaller, the match quality goes higher, and vice versa.

As mentioned, \ts has been used for NLP tasks to infer continuous values for instances.
However, it is important to note that the support of a Gaussian distribution is unbounded, namely $\mathbb{R} = (-\infty, \infty)$.
This does not satisfy the property of absolute zero of scalar annotation in the level of measurement (\S\ref{sec:direct}).
It becomes problematic when it comes to annotating a scalar (continuous) value for extremes such as extremely positive or negative sentiments. 
We address this issue by proposing a novel variant of \ts in the next section.

\subsection{Bounded Variant}
\label{sec:bounded}
\ts can induce a continuous spectrum of instances (such as skill level of game players) by assuming that each instance is represented as a Gaussian distribution.
However, the Gaussian distribution has unbounded support, namely $\mathbb{R} = (-\infty, \infty)$, which does not satisfy the property of \emph{absolute} bounds for appropriate scalar annotation (i.e., ratio scale in the level of measurement).

Thus, we propose a variant of \ts by changing the latent distribution from a Gaussian to a beta,
using a heuristic algorithm based on \ts for inference.
The Beta distribution has natural $[0, 1]$ upper and lower bounds and a simple parameterization:  $S_i \sim  \B_i(\alpha_i, \beta_i)$.  
We choose the scalar response as the mode $\mm[\object]$ of the distribution and the variance as uncertainty:\footnote{We may have instead used the mean ($\mathbb{E}[S_i]=\frac{\alpha_i}{\alpha_i + \beta_i}$) of the distribution, where in a beta  ($\alpha, \beta>1$) the mean is always closer to 0.5 than the mode, whereas mean and mode are always the same in a Gaussian distribution.  The mode was selected owing to better performance in development.}
\vspace{-0mm}
\begin{align}
\mm_i &= \frac{\alpha_i-1}{\alpha_i + \beta_i -2} \label{eq:betamm}\\
\text{Var}_i &= \sigma^2_i = \frac{\alpha_i\beta_i}{(\alpha_i + \beta_i)^2(\alpha_i+\beta_i+1)} \label{eq:betavar}
\end{align}


As in \ts, we iteratively update parameters of instances $\B(\alpha, \beta)$ according to each observation and how it is surprising. 
Similarly to Eqns.~\eqref{eq:tsupdate1} and \eqref{eq:tsupdate2}, we choose the update functions as follows;\footnote{There may be other potential update (and surprisal) functions such as $-\log p$, instead of $1-p$. As in our use of the mode rather than mean as scalar response, we empirically developed our update functions with respect to  annotation efficiency observed through experimentation (\S~\ref{sec:experiments}).}
first, in case that an annotator judged that $S_i$ is preferred to $S_j$ ($S_i \succ S_j$),
\vspace{-0mm}
\begin{align}
& \alpha_i = \alpha_i + \frac{\sigma^2_i}{c} \cdot (1-p_{i \succ j}) \label{eq:betaUpdate1}\\
& \beta_j = \beta_j + \frac{\sigma^2_j}{c} \cdot (1-p_{j \prec i})
\end{align}
in case of ties with $|D|>\epsilon$ and $\mm_i > \mm_j$,
\vspace{-0mm}
\begin{align} 
& \alpha_j = \alpha_j + \frac{\sigma^2_j}{c} \cdot (1-p_{i \equiv j}) \\
& \beta_i = \beta_i + \frac{\sigma^2_i}{c} \cdot (1-p_{i \equiv j})
\end{align}
and in case of ties with $|D|\leqslant\epsilon$, for both $S_i, S_j$,
\vspace{-0mm}
\begin{align} 
& \alpha_{i,j} = \alpha_{i,j} + \frac{\sigma^2_{i,j}}{c} \cdot (1-p_{i \equiv j}) \\
& \beta_{i,j} = \beta_{i,j} + \frac{\sigma^2_{i,j}}{c} \cdot (1-p_{i \equiv j}).\label{eq:betaUpdate6}
\end{align}
Regarding the probability of pairwise comparison between instances, we follow \newcite{Bradley1952} and \newcite{Rao1967} 
to describe the chance of win, tie, or loss, as follows:
\begingroup\makeatletter\def\f@size{9}\check@mathfonts
\def\maketag@@@#1{\hbox{\m@th\large\normalfont#1}}%
\vspace{-0mm}
\begin{align}
& p(S_i \succ S_j) = p(D>\epsilon) = \frac{\pi_i}{\pi_i + \theta\pi_j} \label{eq:rao1} \\
& p(S_i \prec S_j) = p(D<-\epsilon) = \frac{\pi_j}{\theta\pi_i + \pi_j} \label{eq:rao2} \\
& p(S_i \equiv S_j) =  p(|D|\leqslant\epsilon) = \frac{(\theta^2 - 1)\pi_i\pi_j}{(\pi_i + \theta\pi_j)(\theta\pi_i + \pi_j)} \label{eq:rao3} 
\end{align}
\endgroup
where $D=\mm_i - \mm_j$, $\epsilon \geqslant 0$ is a parameter to specify the tie rate,  $\theta=\exp{(\epsilon)}$, and $\pi$ is an exponential score function of $S$; $\pi_i = \exp(\mm_i)$.
\begin{figure}[t]
  \centering
  \SetFigLayout{1}{2}
    \subfigure[$1-p_{i \succ j}$]{\includegraphics[width=38mm]{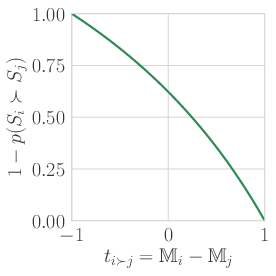}}
    \hfill
    \subfigure[$1-p_{i \equiv j}$]{\includegraphics[width=38mm]{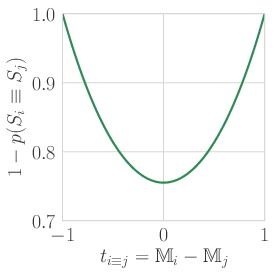}}
	\caption{Surprisal of the outcome for the bounded variant ($\epsilon=0.5)$.}
    \label{fig:updatebeta}
\end{figure}

It is important to note that $\alpha$ and $\beta$ never decrease (because $1-p \geq 0$ as shown Figure~\ref{fig:updatebeta}), which satisfies the property that variance (uncertainty) always decreases as we observe more judgments, as seen in \ts (\S\ref{sec:unbounded}). 
In addition, we do not need individual update functions for $\mu$ and $\sigma^2$, since the mode and variance in beta distribution depend on two shared parameters $\alpha, \beta$ (Eqns.~\ref{eq:betamm} and \ref{eq:betavar}).

Regarding match quality, we use the same formulation as the \ts (Eqn.~\ref{eq:tsmatch}), except that the bounded model uses $\mm$ instead of $\mu$:
\begingroup\makeatletter\def\f@size{10}\check@mathfonts
\def\maketag@@@#1{\hbox{\m@th\large\normalfont#1}}%
\vspace{-0mm}
\begin{align}
q(\gamma, S_i, S_j) =  \sqrt[]{\frac{2\gamma^2}{c^2}} \exp{\left(- \frac{(\mm_i-\mm_j)^2}{2 c^2}  \right) } \label{eq:match} 
\end{align}
\endgroup

\section{\methodfull}
\label{sec:method}
\begin{figure}[t]
	\centering
	\includegraphics[width=55mm]{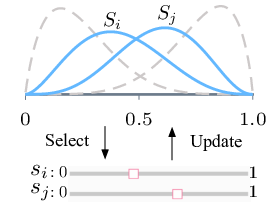}
    \vspace{-0mm}
	\caption{Illustrative example of the EASL protocol. Each instance is represented as a beta distribution. Instances are chosen to annotate according to the variance and match quality, and the parameters are updated iteratively.}
    \label{fig:easl}
	\vspace{-0mm}
\end{figure}
In the previous section, we propose a \emph{bounded} online ranking aggregation model for scalar annotation.
However, the amount of update by a pairwise judgment depends only on the distance between instances, not on the distance from the bounds (i.e., 0 and 1).
To integrate this property into the online ranking aggregation model, we propose \method (\methodfull) that combines benefits from both direct assessment (\direct) and bounded online ranking aggregation model (\rank).\footnote{~\newcite{2018arXiv180305928N} recently proposed a similar approach named RankME, which is a variant of \direct with comparing multiple instances at a time. It can also be regarded as a batch-learning variant of \method without probabilistic parameterization.}

Similarly to \rank, \method parameterizes each instance by a beta distribution (Eqns.~\ref{eq:betamm} and ~\ref{eq:betavar}), and the parameters are inferred using a computationally efficient and easy-to-implement heuristic.
The difference from \rank is the type of annotation. 
While we ask for discrete pairwise judgment ($\succ,\prec,\equiv$) between $S_i$ and $S_j$ in \rank, here we directly ask for scalar values for them (denoted as $s_i$ and $s_j$) as in \direct.
Thus, given an annotated score $s_i$ which is normalized between [0,1], we change the update functions as follows:
\vspace{-0mm}
\begin{align}
\alpha_i &= \alpha_i + s_i \label{eq:updatealpha}\\
\beta_i &= \beta_i + (1 - s_i) \label{eq:updatebeta}
\end{align}

This procedure may look similar to \direct, where $s_i$ is simply accumulated and averaged at the end.
However, there are two differences. 
First, as illustrated in Figure~\ref{fig:easl}, \method parameterizes each instance as a probability distribution while \direct does not.
Second, \direct  elicits annotations independently per element, whereas \method elicits annotations on elements in the context of other elements selected jointly according to match quality.

Further, \direct generally uses a batch style annotation scheme, where the number of annotations per instance is independent from the latent scalar values.  On the other hand, \method uses online learning, which impacts the calculation of match quality. 
This allows us to choose instances to annotate by order of uncertainty for each instance, and  as in \rank, the match quality (Eqn.~\ref{eq:match}) enables us to consider similar instances in the same context.

\section{Experiments}
\vspace{-0mm}
\label{sec:experiments}
To compare different  annotation methods, we conduct three experiments: (1) lexical frequency inference, (2) political spectrum inference, and (3) human evaluation for machine translation systems.

In all experiments, data collection is conducted through Amazon Mechanical Turk (AMT).
We ask annotators who meet the following minimum requirements:\footnote{In all experiments, we set the reward of single instance to be \$0.01 (i.e., \$0.05 in \rank and \method). This is \$8/hour, assuming that annotating one instance takes five seconds. Prior to annotation, we run a pilot to make sure that the participants understand the task correctly and the instructions are clear.} living in the US, overall approval rate $>$ 98\%, and number of tasks approved $>$ 500.

The experimental setting for \direct is straightforward.
We ask annotators to annotate a scalar value for each instance, one item at a time.
We collect ten annotations for each instance to see the relation between the number of annotations and accuracy (i.e., correlation).

\begin{figure}[t]
	\centering
	\includegraphics[width=70mm]{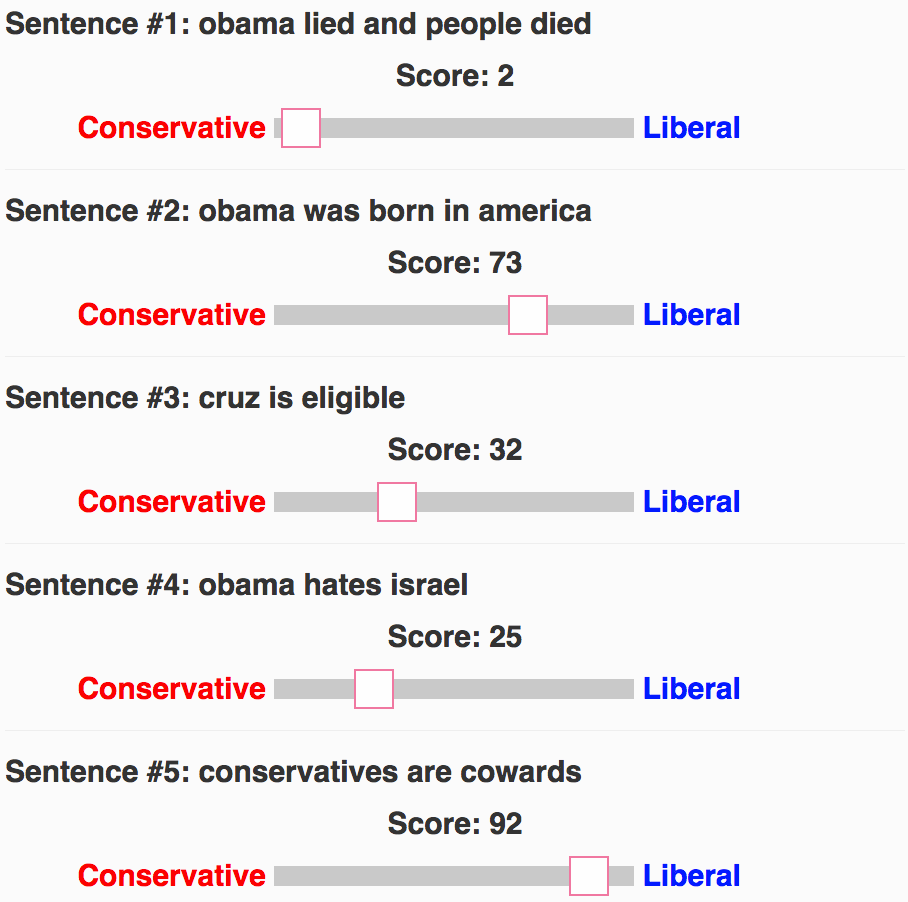}
    \vspace{-0mm}
	\caption{Example of partial ranking with scalars (HITS)}
    \label{fig:partialrank}
	\vspace{-0mm}
\end{figure}
To set up the online update in  \rank and \method,
we use a \emph{partial ranking} framework with scalars, where annotators are asked to rank and score $n$ instances at one time as illustrated in Figure~\ref{fig:partialrank}.
In all three experiments, we fix $n=5$.
The partial ranking yields $\binom{n}{2}$ pairwise comparisons for \rank and $n$ scalar values for \method.\footnote{The partial ranking can be regarded as mini-batching.}
It is important to note that we can simultaneously retrieve pairwise judgments ($\succ,\prec,\equiv$) as well as scalar values from this format.

\begin{algorithm}[t]
\small
\DontPrintSemicolon
\SetAlgoNoEnd
\KwIn{Instances $\{S_1^N\}$}
\KwOut{Updated instances $\{S_1^N\}$}
\tcc{\small Initialize params}
$(\alpha_i, \beta_i)_{\in S}  = (\alpha_i^{\text{init}}, \beta_i^{\text{init}})$\\
\tcc{\small Update $S$ over iterations}
\ForEach{iteration}{
	HITS = SampleByMatchQuality($S, N, n$)\\
    $A$ = Annotate(HITS)\\
    \For(\tcp*[f]{\small Update $S$}){obs $\in A$}{ 	
    	$i, j, d$ = parseObservation($obs$)\\
        $\alpha_{i,j}, \beta_{i,j}$ = update($i,j,d$)\\
    }
}
\KwRet{$S$}\\
\SetKwProg{Fn}{Function}{}{}
\Fn{SampleByMatchQuality($S, N, n$)}{
	$k = N/n$\\
	descendingSort($S$, key=Var[$S$])\\
 	$S'=$ top-k instances of $S$\\
    HITS = []\\
    \ForEach{$S_i \in S'$}{
    	m = []\\
    	\ForEach{$S_j \in S_{/S'}$}{
        	m.append([matchQuality($S_i, S_j$), $j$])\\
        }
        $p$ = normalize(m)\\
		$\tilde{S}$ = sampling n-1 items by $p$\\
        HITS.append([$S_i, \tilde{S}$])\\
    }
	\KwRet{HITS}\\
}
\caption{Online pairwise ranking aggregation with bounded support.}
\label{alg:beta}
\end{algorithm}

In each iteration, $n$ instances are selected by variance and match quality.
We first select top k ($=N/n$) instances according to the variance, and for each selected instance we choose the other $n-1$ instances to be compared based on match quality.
This approach has been used in the NLP community in tasks such as for assessing machine translation quality \cite{bojar-EtAl:2014:W14-33,sakaguchi-post-vandurme:2014:W14-33,bojar-EtAl:2015:WMT,bojar-EtAl:2016:WMT1} to collect pairwise judgments efficiently.
The detailed procedure of iterative parameter updates in the \rank and \method is described in 
Algorithm~\ref{alg:beta}.
As mentioned in Section~\ref{sec:method}, the main difference between \rank and \method is the update functions (line 7).

Model hyper-parameters in \rank and \method are set as follows; each instance is initialized as $\alpha_i^{\text{init}}=1.0$, $\beta_i^{\text{init}}=1.0$.
The skill chain parameter $\gamma$ and tie-rate parameter $\epsilon$ are set to be 0.1.\footnote{We explored the hyper-parameters $\gamma, \epsilon$ in a pilot task.}

\subsection{Lexical Frequency Inference}
\label{sec:lexical}
\vspace{-0mm}
In the first experiment, we compare the three scalar annotation approaches on lexical frequency inference, in which we ask annotators to judge frequency (from very rare to very frequent) of verbs that are randomly selected from the corpus of Contemporary American English (COCA)\footnote{\url{https://www.wordfrequency.info/}}.  We include this task for evaluation owing to its non-subjective ground truth (relative corpus frequency) which can be used as an oracle response we would like to maximally correlate with.\footnote{Lexical frequency inference is  an established experiment in (computational) psycholinguistics. E.g., human behavioral measures have been compared with  predictability and bias in various  corpora \cite{balota-EtAl:1999,fine-EtAl:2014:P14-2}.}
\begin{figure}[t]
	\centering
    \vspace{-0mm}
	\includegraphics[width=77mm]{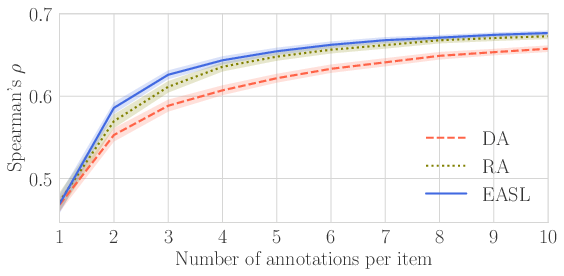}
	\includegraphics[width=77mm]{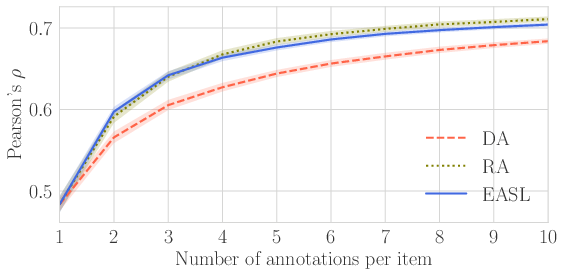}
	\caption{Spearman's (top) and Pearson's (bottom) correlations with three difference methods on lexical frequency inference annotation: direct assessment (\direct), online ranking aggregation (\rank), and \method. The shade for each line indicates 95\% confidence intervals by bootstrap resampling (running 100 times).}
    \label{fig:results_vocab_corr}
	\vspace{-0mm}
\end{figure}
\begin{figure}[t]
	\centering
    \vspace{-0mm}
	\includegraphics[width=77mm]{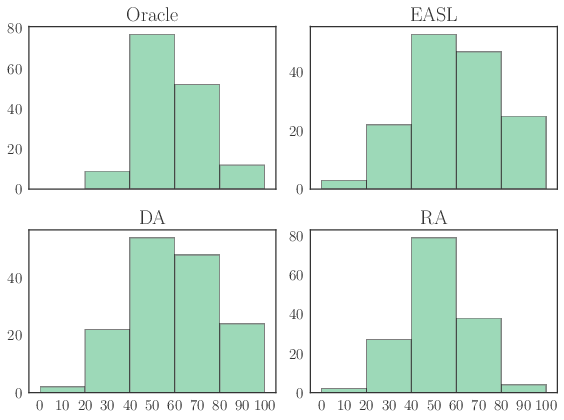}
	\caption{Histograms of scalar values on lexical frequency obtained by each annotation scheme (direct assessment (\direct), online ranking aggregation (\rank), and \method), and the oracle. The scalar annotations are put into five bins to see the overall distribution. The scalar in the oracle is normalized as $log_{10}$(frequency($S_i$)) / $\max log_{10}$(frequency($S$)).}
    \label{fig:result_vocab_hist}
	\vspace{-0mm}
\end{figure}

We randomly select 150 verbs from COCA; the log frequency ($\log_{10}$) is regarded as the oracle.
In \direct, each instance is annotated by 10 different annotators.\footnote{The agreement rate in \direct (10 annotators) is 0.37 in Spearman's $\rho$. Considering the difficulty of ranking 150 verbs, this rate is fair.}
In the \rank and \method, annotators are asked to rank/score five verbs for each HIT ($n=5$).
Each iteration contains 20 HITS and we run 10 iterations, which means that total number of annotations is the same in \direct, \rank, and \method.\footnote{Technically, the number of annotations per instance vary in \rank and \method, because they choose instances by match quality at each iteration.}
\begin{figure}[t]
  \centering
  \SetFigLayout{1}{4}
    \subfigure[Iter 0]{\includegraphics[width=18mm]{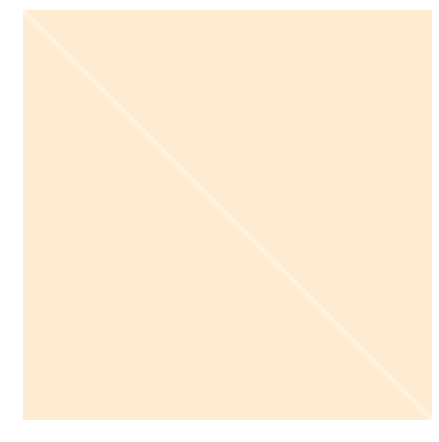}}
    \hfill
    \subfigure[Iter 3]{\includegraphics[width=18mm]{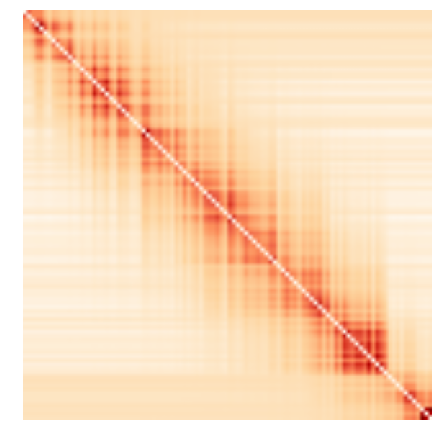}}
    \hfill
    \subfigure[Iter 6]{\includegraphics[width=18mm]{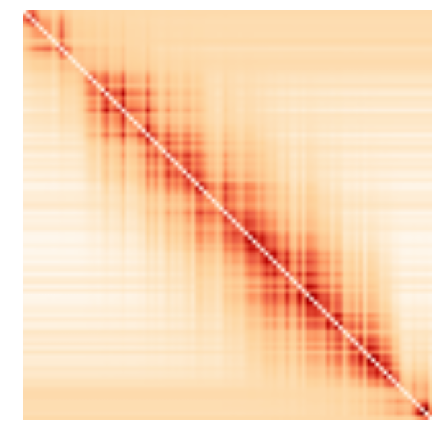}}
    \hfill
    \subfigure[Iter 9]{\includegraphics[width=18mm]{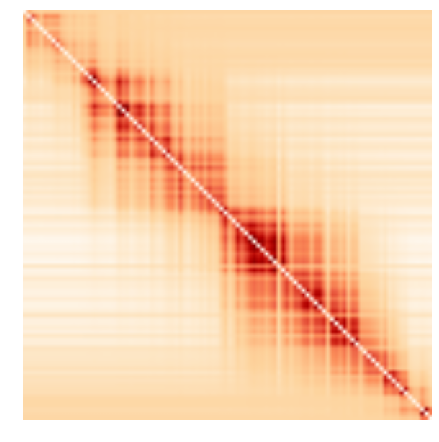}}
    \vspace{-0mm}
	\caption{Heatmaps of match quality distribution across the cross-product of instances ordered by the oracle (i.e., $log_{10}$(frequency)).}
    \label{fig:match-quality}
	\vspace{-0mm}
\end{figure}

Figure \ref{fig:results_vocab_corr} presents Spearman's and Pearson's correlations, indicating how accurately each annotation method obtains scalar values for each instance.
Overall, in all three methods, the correlations are increased as more annotations are made.
The result also shows that \rank and \method approaches achieve high correlation more efficiently than \direct. 
The gain of efficiency from \direct to \method is about 50\%; two iterations in \method achieves a close Spearman's $\rho$ to three annotators in \direct. 

Figure \ref{fig:result_vocab_hist} presents the results of the final scalar values that each method annotated.
The distribution of the histograms shows that overall three methods successfully capture the latent distribution of scalar values in the data. 

Figure~\ref{fig:match-quality} shows a dynamic change of match quality.
In the beginning (iteration 0), all the instances are equally competitive because we have no information about them and initialize them with the same parameters.
As iterations go on, the instances along the diagonal have higher match quality, indicating that competitive matches are more likely to be selected for a next iteration.
In other words, match-quality helps to choose informative pairs to compare at each iteration, which reduces the number of less informative annotations (e.g., a pairwise comparison between the highest and lowest instances).

\subsection{Political Spectrum Inference}
\label{sec:political}
\vspace{-0mm}
\begin{figure}[t]
	\centering
	\includegraphics[width=77mm]{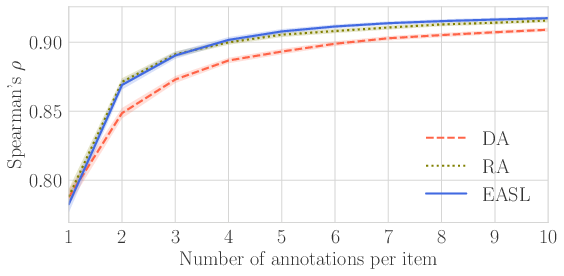}
	\includegraphics[width=77mm]{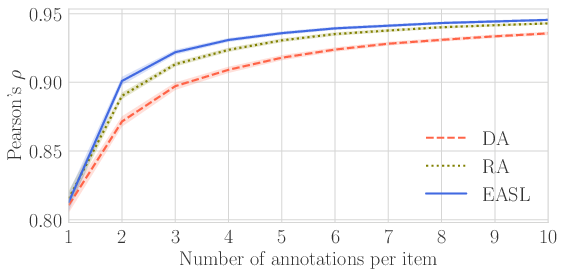}
	\caption{Spearman's (top) and Pearson's (bottom) correlations with three difference methods on political spectrum annotation: direct assessment (\direct), online ranking aggregation (\rank), and \method }
    \label{fig:results_political_corr}
	\vspace{-0mm}
\end{figure}

In the second experiment, we compare the three scalar annotation methods for political spectrum inference.
We use the Fine-Grained Political Statements dataset \cite{bamman-smith:2015:EMNLP}, 
which consists of 
766 propositions collected from political blog comments, paired with judgments about the political belief of the statement (or the person who would say it)
based on the five ordinals: {\em very conservative} (-2), {\em slightly conservative} (-1), {\em neutral} (0), {\em slightly liberal} (1), and {\em very liberal} (2).
We normalize the ordinal scores between 0 and 1.
The dataset contains the mean scores by aggregating 7 annotations for each proposition.\footnote{We stress that the oracle here derives from subjective annotations: it does not necessarily reflect the true latent scalar values for each instance. However, in this experiment, we use them as a tentative oracle to compare three scalar annotation methods objectively.}

\begin{figure}[t]
	\centering
    \vspace{-0mm}
	\includegraphics[width=77mm]{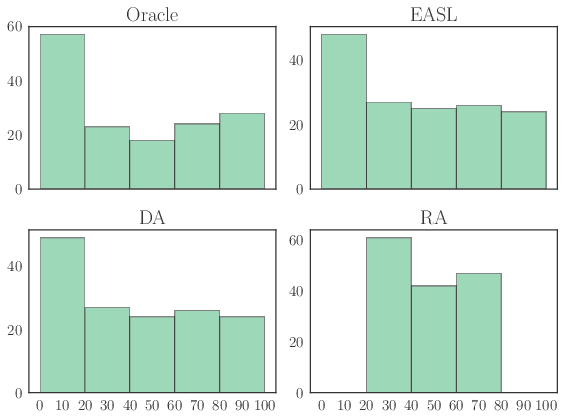}
	\caption{Histograms of scalar values on political spectrum obtained by each annotation scheme (\direct, \rank, \method) and the oracle. Scalars are put into five bins to see the overall distribution.}
    \label{fig:result_political_hist}
	\vspace{-0mm}
\end{figure}

We randomly choose 150 political propositions from the dataset (see the histogram in Figure~\ref{fig:result_political_hist} oracle).\footnote{The agreement rate in \direct (among 10 annotators) is 0.67 in Spearman's $\rho$. This is significantly high, considering the difficulty of ranking 150 instances in order.}
The experimental setting (i.e., the number of annotations per instance, the number of iterations, and the number of HITS in each iteration) is the same as the lexical frequency inference experiment (\S\ref{sec:lexical}).

Figure \ref{fig:results_political_corr} shows Spearman's and Pearson's correlations to the oracle by each method.
Overall, all the three methods achieve strong correlation above 0.9.
We also find that \rank and \method reach high correlation more efficiently than \direct as in the lexical frequency inference experiment (\S\ref{sec:lexical}).
The gain of efficiency from \direct to \method is about 50\%; 4-way redundant annotation in \method achieves a close Spearman's $\rho$ to 6-way redundancy in \direct. 
\begin{table}[t]
\centering
\small
\begin{tabular}{|lrrrr|}
\hline
\scriptsize{Propositions}      & \scriptsize{Gold} & \scriptsize{\direct} & \scriptsize{\rank} & \scriptsize{\method}   \\ \hline\hline
the republicans are useless    & 100    & 91.7  & 75.8  & 91.9 \\
obama is right                 & 92.9   & 90.1  & 74.6  & 90.0 \\ \hdashline
hillary will win               & 78.6   & 86.3  & 72.9  & 86.4 \\
aca is a success               & 75.0   & 78.2  & 68.3  & 77.3 \\ \hdashline
harry reid is a democrat       & 53.6   & 55.5  & 55.8  & 55.9 \\
ebola is a virus               & 50.0   & 53.0  & 53.8  & 53.5 \\ \hdashline
cruz is eligible               & 32.2   & 31.0  & 44.0  & 31.4 \\
global warming is a religion   & 28.6   & 22.4  & 37.3  & 23.0 \\ \hdashline
bush kept us safe              & 10.7   & 9.6   & 31.5  & 9.6  \\
democrats are corrupt          & 0.0    & 7.1   & 29.9  & 7.4  \\ \hline
\end{tabular}
\caption{Example propositions and the scalar political spectrum ranged between 0 (\emph{very conservative}) and 100 (\emph{very liberal}) by each approach: direct assessment, online ranking aggregation, and \method. The dashed lines indicate a split by 5-ary ordinal scale.}
\label{tab:political}
\vspace{-0mm}
\end{table}

Figure \ref{fig:result_political_hist} presents the results of the annotated scalar values by each method.
The distribution of the histograms shows that \direct and \method successfully fit to the distribution in the oracle, whereas \rank converges to a rather narrow range.
This is because of the ``lack of distance from bounds'' in \rank that is explained in \S\ref{sec:method}.
We note that renormalizing the distribution in \rank will not address the issue. 
For instance, when the dataset has only liberal propositions, \rank still fails to capture the latent distribution because it looks only at relative distances between instances but not the distance from bounds.
Table~\ref{tab:political} shows the examples of scalar annotations by each method.
Again, we see that \rank approach has a narrower range than the oracle, \direct, and \method.

\subsection{Ranking Machine Translation Systems}
\vspace{-0mm}
In the third experiment, we apply the scalar annotation methods for evaluating machine translation systems.
This is different from two previous experiments, because the main purpose is to rank the MT systems ($S_1^N$) rather than the adequacy ($q$) of each MT output for a given source sentence ($m$).
Namely, we want to rank $S_i$ by observing $q_{i,m}$.

We use WMT16 German-English translation dataset~\cite{bojar-EtAl:2016:WMT1}, which consists of 2,999 test set sentences and the translations from 10 different systems with \direct annotation.
Each sentence has its adequacy score annotation between 0 and 100, and the average adequacy scores are computed for each system for ranking.
In this setting, annotators are asked to judge adequacy of system output(s) with the reference being given.
The official scores (made by \direct) and ranking in WMT16 are used as the oracle in this experiment. 

In this experiment, we replicate \direct and run \method to compare the efficiency.
We omit \rank in this experiment, because it does not necessarily capture the distance from bounds as shown in the previous experiment (\S\ref{sec:political}).
In \direct, 33,760 translation outputs (3,376 sentences per system in average) are randomly sampled without replacement to make sure that it reaches up to the same result as oracle when the entire data are used.

In \method, we assume that adequacy ($q$) of an MT output by system ($S_i$) for a given source sentence ($m$) is drawn from beta distribution: $q_{i,m} \sim \mathcal{B}(\alpha_i, \beta_i)$.\footnote{This is the same setting as WMT14, WMT15, and WMT16~\cite{bojar-EtAl:2014:W14-33,bojar-EtAl:2015:WMT}, although they used \ts (Gaussian) instead of \method to rank systems.}
Annotators are asked to judge adequacy of system outputs by scoring 0 and 100.
Similarly to the previous experiments (\S~\ref{sec:lexical} and \S~\ref{sec:political}), we use the partial ranking strategy, where we show $n=5$ system outputs (for the same source sentence $l$) to annotate at a time.
The procedure of parameter updates is the same as previous experiments (Algorithm~\ref{alg:beta}).

\begin{figure}[t]
	\centering
    \vspace{-0mm}
	\includegraphics[width=80mm]{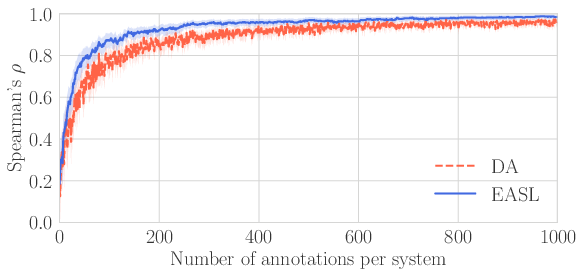}
	\caption{Spearman's correlation on ranking machine translation systems on WMT16 German-English data: direct assessment (\direct), and \method. The shade for each line indicates 95\% confidence intervals by bootstrap resampling (running 100 times).}
    \label{fig:result_MT}
	\vspace{-0mm}
\end{figure}

We compare the correlations (Spearman's $\rho$) of system ranking with respect to the number of annotations per system, and the result is shown in Figure~\ref{fig:result_MT}.
As seen in the previous two experiments, \method achieves higher Spearman’s correlation on ranking MT systems with smaller number of annotations than the baseline method (\direct), which means \method is able to collect annotation more efficiently.
The result shows that \method can be applied for efficient system evaluation in addition to data curation.

\section{Conclusions}

We have presented an efficient, online model to elicit scalar annotations for computational linguistic datasets and system evaluations.
The model combines two approaches for scalar annotation: direct assessment and online pairwise ranking aggregation. 
We  conducted three illustrative experiments on lexical frequency inference, political spectrum inference, and ranking machine translation systems.
We have shown that our approach, \method (\methodfull), outperforms direct assessment in terms of annotation efficiency and outperforms online ranking aggregation in terms of accurately capturing the latent distributions of scalar values.
The significant gains demonstrated suggests \method as a promising approach for future dataset curation  and system evaluation in the community.



\section*{Acknowledgments}
We are grateful to Rachel Rudinger, Adam Teichert, Chandler May, Tongfei Chen, Pushpendre Rastogi, and anonymous reviewers for their useful feedback. This work was supported in part by IARPA MATERIAL and DARPA LORELEI. 
The U.S. Government is authorized to reproduce and distribute reprints for Governmental purposes. 
The views and conclusions contained in this publication are those of the authors and should not be interpreted as representing official policies of the U.S. Government.

\bibliographystyle{acl_natbib}
\bibliography{acl2018}

\end{document}